\documentclass[sigconf, nonacm]{acmart}

\AtBeginDocument{%
  }

\usepackage{amsmath}
\usepackage{amsfonts}
\usepackage{multirow}
\usepackage{cleveref}
\usepackage{booktabs}
\usepackage{tabularx}
\usepackage{makecell}
\usepackage{dcolumn}
\usepackage{graphicx}
\crefname{table}{Table}{Table}
\crefname{figure}{Figure}{Figure}

\usepackage{newfloat}
\usepackage{listings}

\begin{document}

\title{Evolving in Tasks: Empowering the Multi-modality Large Language Model as the Computer Use Agent}





\author{Yuhao Cheng}
\email{yuhao.cheng@outlook.com}
\affiliation{%
  \institution{Lenovo Research}
  \state{Beijing}
  \country{China}
}
\author{Liang Tang}
\email{tangliang5@lenovo.com}
\affiliation{%
  \institution{Lenovo Research}
  \state{Beijing}
  \country{China}
}
\author{Shuxian Li}
\email{lisx14@lenovo.com}
\affiliation{%
  \institution{Lenovo Research}
  \state{Beijing}
  \country{China}
}

\author{Yukang Huo}
\email{yukanghuo.ai@gmail.com}
\affiliation{%
  \institution{China Agricultural University}
  \state{Beijing}
  \country{China}
}

\author{Tiaonan Duan}
\email{duantn1@lenovo.com}
\affiliation{%
  \institution{Lenovo Research}
  \state{Beijing}
  \country{China}
}

\author{Kaer Huang}
\email{huangke1@lenovo.com}
\affiliation{%
  \institution{Lenovo Research}
  \state{Beijing}
  \country{China}
}

\author{Yanzhe Jing}
\email{jingyz1@lenovo.com}
\affiliation{%
  \institution{Lenovo Research}
  \state{Beijing}
  \country{China}
}

\author{Yiqiang Yan}
\email{yanyq@lenovo.com}
\affiliation{%
  \institution{Lenovo Research}
  \state{Beijing}
  \country{China}
}








\renewcommand{\shortauthors}{Yuhao et al.}

\begin{abstract}
Computer use agents represent an emerging area in artificial intelligence, aiming to operate computers autonomously to fulfill user tasks, attracting significant attention from both industry and academia. However, the performance of existing agents remains insufficient for practical deployment. In this paper, we propose the \textit{Self-Evolution Agent (SEA)} for computer operation, alongside three core innovations in data generation, reinforcement learning, and model enhancement to develop this agent. Specifically, we first design an automatic pipeline to generate verifiable task trajectories for training. Second, we propose \textit{Efficient Step-wise Reinforcement Learning} to reduce the substantial computational overhead of long-horizon training. Finally, we introduce a model enhancement method that integrates grounding and planning capabilities into a single model without additional training. Leveraging these innovations, our SEA (with only 7B parameters) outperforms existing models of the same parameter scale and achieves performance comparable to larger models (e.g., 32B/72B parameters) on computer use tasks. We plan to release the model weights and related code as open-source resources in the future.
\end{abstract}


\maketitle

\section{Introduction}
Computer-use agents have been an attractive technology that follows users' instructions to manipulate computers and match their expectations, marking a milestone towards general artificial intelligence. 
As the browser is the most important software on a computer, the agent's functions are divided into two categories. The first one involves manipulating the web browser, and the second consists of using other software, such as office applications, media players, IDEs, and so on. 
To achieve the goal of computer use, the agent must understand the computer environment, provide a plan based on the instructions, and ultimately manipulate the computer. For the computer environment, researchers will take a screenshot as a representation of that. With the development of Multi-modal Large Language Models (MLLMs), such as~\cite{bai2025qwen2,guo2025seed1}, they possess a remarkable ability for perception and reasoning, which can meet the requirements of a computer-use agent. Meanwhile, these models satisfy the requirement of making execution plans and operating the computer after some post-training.
\begin{figure}
    \centering
    \includegraphics[width=\linewidth]{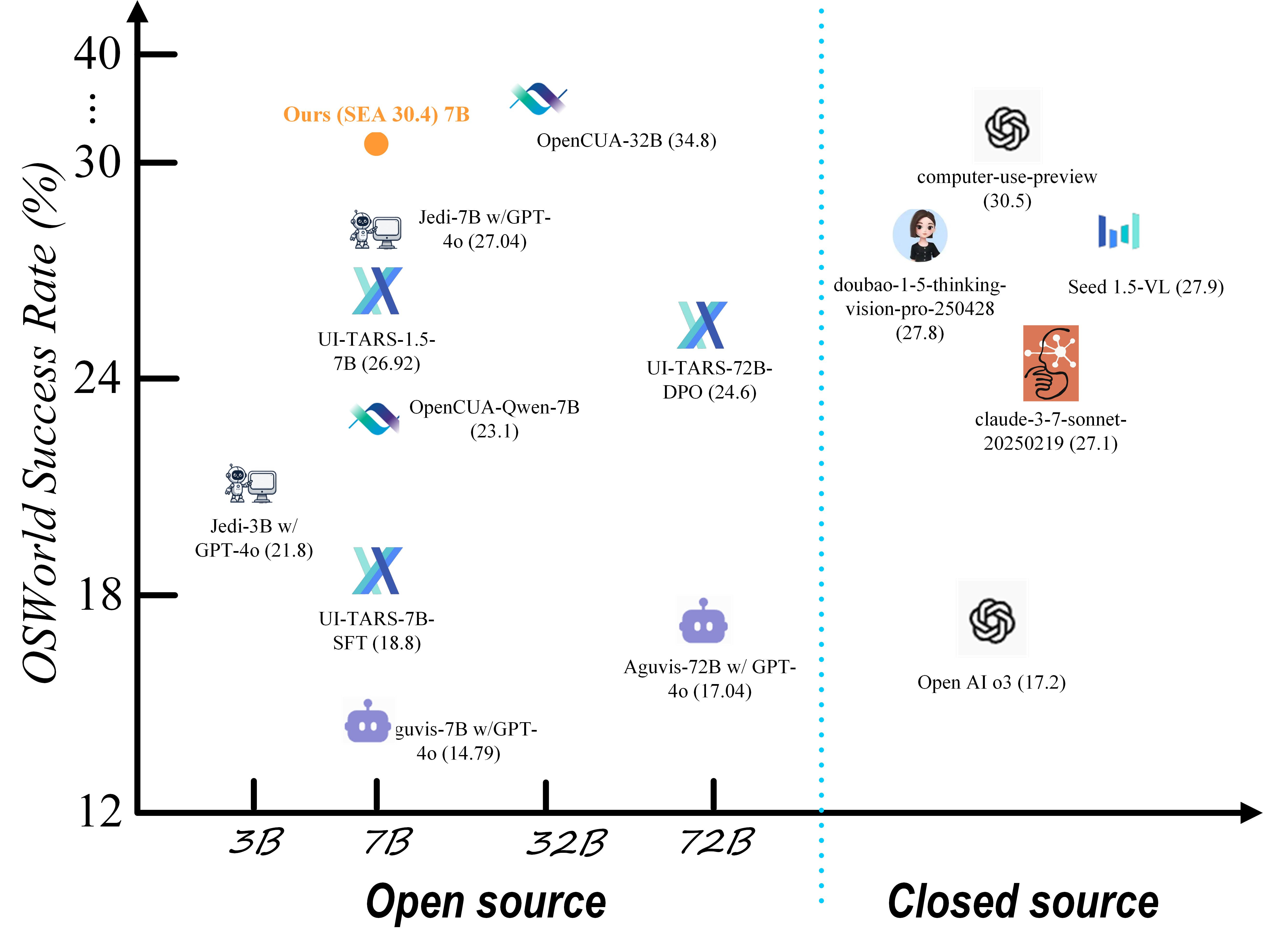}
    \caption{The comparison with the SOTA methods. This scatter plot depicts the OSWorld success rates of different open-source and closed-source models. Our proposed model achieves the highest success rate among open-source 7B models and substantially outperforms some other open-source alternatives with a larger number of parameters, even coming close to the performance of certain closed-source models.}
    \label{fig:comparison}
    \vspace{-5mm}
\end{figure}
Despite significant progress in the field of intelligent agents, building robust agents for real-world deployment remains highly challenging, particularly in scenarios that involve complex multimodal reasoning and dynamic, uncertain environments.
For example, when operating websites, there are always some pop-up advertisements, and their contents vary from time to time. 
One of the most critical challenges in training intelligent agents is acquiring high-quality data. Current mainstream data construction approaches predominantly rely on manually designed task templates and annotations. This process is not only costly and difficult to scale but also often fails to meet the semantic coverage and diversity required by complex tasks. Moreover, real-world tasks typically involve multi-step, long-horizon interactions, where informative reward signals only emerge after completing the entire task or passing through several key steps. 
Such sparse reward settings make it difficult for the agent to establish a clear correlation between individual actions and final outcomes, leading to suboptimal task performance. However, when handling complex GUI tasks, encoding image data often requires processing a large sequence of tokens, many of which contribute minimally to task execution and reasoning but incur significant computational costs. In addition, existing reasoning mechanisms frequently depend on external models to assist in evaluating and selecting trajectories. This reliance introduces significant computational overhead and increases system complexity, further limiting the efficiency and scalability of the overall agent architecture.

To address these challenges, we propose a set of novel agent training methods, designed to enable efficient and scalable task learning and execution. In the data acquisition phase, we propose a closed-loop pipeline to generate the verifiable tasks, which leverages a task-generation agent and a code-generation agent to automatically synthesize tasks along with their corresponding execution and verification Python programs. 
This dual-code generation strategy ensures that each task is accompanied by a well-defined success criterion and a programmatic validation method, effectively forming a closed-loop supervision structure. By tightly coupling the task definition with its success condition, the proposed pipeline significantly enhances both the effectiveness and usability of the generated data.
Furthermore, we use a virtual environment to run each task using its corresponding execution program and apply the verification program to retain only those tasks that are successfully completed. 

And then we propose the Generation and Assessment for Trajectory Extraction (GATE) method, designed to construct high-quality and representative training trajectories that improve agent learning efficiency. Specifically, for each validated task, we perform multiple rounds of inference to generate diverse candidate trajectories. 
As training progresses, we gradually replace the inference model with our agent during the middle and later stages of training. Each generated trajectory is then filtered again using the verification code to ensure successful task completion. Among the successful trajectories for the same task, we prioritize those with the fewest required steps, selecting more efficient execution paths. Additionally, we train a step filtering model to eliminate redundant or incorrect steps within each trajectory, resulting in high-quality, concise training data.

Another challenge is how to efficiently use the sparse rewards, so during the training phase, we introduce the Trajectory Reasoning by Step-wise Reinforcement Learning(TR-SRL). This component evaluates each step in a task trajectory by comparing the agent’s execution results against the corresponding ground-truth outcomes. In addition, TR-SRL assesses the consistency between the agent’s generated thoughts and executed actions at each step, ensuring alignment between internal reasoning and external behavior. 

At last, we propose a grounding-based enhancement method to further improve the generalization ability and effect of the model. Specifically, we first train a grounding model and integrate its weights, which effectively improves the performance of both grounding and planning.
In addition, we introduce a Temporal Compressed Sensing Mechanism to optimize perception efficiency by modeling the importance of input image tokens over time. By retaining only the most essential tokens, the model achieves a significant reduction in training cost while maintaining critical semantic content and improving inference efficiency. 

Our main contributions can be summarized as follows:
\begin{itemize}
    \item We propose a data generation pipeline to produce the closed-loop verifiable tasks' trajectory for reinforcement learning. The proposed pipeline will simultaneously generate the task instruction, execution program, and verification program. To further refine data quality, we introduce the GATE, which identifies optimal training steps via trajectory sampling and evaluation.
    \item We propose the TR-SRL(Trajectory Reasoning by Step-wise Reinforcement Learning), which uses the step-wise training manner to replace the long-horizon training of manipulating the computer.  
    \item We proposed a Grounding-Based Generalization Enhancement method to combine the planning ability and grounding ability into a unified model. The proposed method not only can keep the original models' performance but also can improve it. 
\end{itemize}
In summary, based on these proposed methods, including verifiable data generation, efficient reinforcement learning, and model enhancement, we create the Self-Evolution Agent (SEA) for computer use. This agent utilizes the synthesis of verifiable tasks to evolve through step-wise reinforcement learning. As shown in ~\cref{fig:comparison}, our agent outperforms the 7B models and achieves comparable performance despite having a larger number of parameters. 
\section{Related Works}
\subsection{High-quality Training Data}
High-quality data is essential for bridging the gap between the symbolic world~\cite{xu2023symbol} and the digital world~\cite{wu2024copilot}, thereby advancing the development of intelligent agents. Early works such as Rico~\cite{deka2017rico} and Mini-Wob~\cite{shi2017world} provided serialized GUI data and low-level keyboard/mouse interactions for mobile applications and web-based tasks, respectively. Subsequent studies have further expanded data resources for mobile~\cite{rawles2023androidinthewild, zhang2024android, lu2024gui, chai2024amex}, web~\cite{liu2018reinforcement, lu2024weblinx, murty2024nnetscape}, and desktop applications~\cite{chen2024guicourse}. To train agents more efficiently, researchers typically use trajectory data, which consists of sequences containing GUI information, including both low-level and high-level instructions and corresponding operations~\cite{li2024effects, zhang2024agentohana, zheng2024agentstudio}. In addition, some studies utilize demonstration data, filtered imitation learning, and self-collected data~\cite{pan2024autonomous, lai2024autowebglm} to fine-tune vision-language models (VLMs), such as AutoUI and CogAgent~\cite{kapoor2024omniact, zhang2023you}. PAE~\cite{zhou2025proposer} generates tasks automatically to provide diverse training data for agents.

However, these methods face several challenges in practical applications: reliance on costly manually annotated task templates, inability of static data to adapt to real-world dynamics, and inconsistent quality of task trajectories—all of which severely limit the effectiveness and adaptability of agents in diverse environments. To address these limitations, in this paper, we propose a closed-loop pipeline to generate high-quality, verifiable tasks with correct execution trajectories, ensuring both data scalability and reliability.
\subsection{Browser and Computer Use Agent}
Recent advancements in multi-modal large language models (MLLMs) have empowered agents to tackle increasingly complex tasks within graphical user interfaces (GUIs). Models such as SeeClick~\cite{cheng2024seeclick} and ScreenAgent~\cite{niu2024screenagent} utilize vision-language models (VLMs) for interpreting UI screens and executing interactive tasks based on multimodal prompts. GUI-R1~\cite{luo2025gui} further enhances UI grounding performance by integrating reinforcement learning (RL) mechanisms into the agent architecture.

For web-centric scenarios, Song~\cite{song2024beyond} proposes an API-driven web agent framework that incorporates task-specific background knowledge to execute complex network operations. Distinct from API-reliant counterparts, modern browser agents emphasize human-like interactive capabilities: WebSailor~\cite{li2025websailor}, an open-source browser agent, combines multi-modal web perception (HTML structure + visual elements) with the DUPO RL algorithm to handle cross-page tasks and dynamic content loading, achieving state-of-the-art performance on benchmarks like BrowseComp. Complementing these, the AgentWorkflow Memory (AWM) module~\cite{wang2024agent} optimizes memory management by selectively retrieving task-relevant workflows to guide sequential actions. Beyond general-purpose agents, specialized tools like PresentAgent~\cite{shi2025presentagent} have emerged, which specializes in automated presentation generation.

Despite substantial progress in perception, reasoning, and execution, existing approaches—particularly browser agents—face pronounced limitations: (1) Sparse rewards are exacerbated in long-horizon browser tasks (e.g., multi-step e-commerce operations) due to intermittent feedback from dynamic web content; (2) Training costs remain prohibitive for agents handling diverse browser environments (e.g., varying UI layouts and JavaScript rendering); (3) Few methods model the thought-action consistency in action sequences, leading to misalignment between planned operations (e.g., "click checkout") and actual GUI interactions. To address these issues, this paper proposes a step-wise RL training paradigm, replacing monolithic long-horizon training through tailored reward design and advanced RL strategy optimization.
\subsection{Reinforcement Learning}
As the fascinating performance of the DeepSeek-R1~\cite{guo2025deepseek}, researchers have noticed that the great improvement of the Post-training phase in improving the performance of the LLM, especially compared to the minor improvement of the pre-training phase. 
OpenAI’s o1~\cite{jaech2024openai} and DeepSeek-R1~\cite{guo2025deepseek} demonstrate strong capabilities in tasks such as mathematical reasoning~\cite{shao2024deepseekmath}, code generation~\cite{liu2025code}, and multimodal reasoning~\cite{huang2025vision, liu2025seg} through structured reward signals. Sweet-RL~\cite{zhou2025sweet} introduces a multi-round DPO framework to improve long-horizon behavior in language agents, while ARPO~\cite{lu2025arpo} increases success rates on complex multi-turn GUI tasks by incorporating experience replay and task selection strategies. DISTRL~\cite{wang2024distrl} boosts training efficiency and success rates for GUI agents in mobile device control tasks through its novel asynchronous distributed reinforcement learning framework and the custom A-RIDE algorithm, while also enhancing generalization across diverse tasks. UI-TARS~\cite{qin2025ui} significantly improves performance in complex GUI tasks through innovations in enhanced perception, unified action modeling, and dual-stage reasoning. Moreover, some methods propose interactive refinement of LLMs/VLMs—particularly Web/GUI agents—using self-evaluated feedback~\cite{pan2024autonomous, bai2024digirl, putta2024agent, wang2024distrl}.

In our proposed method, we leverage the SOTA RL training method, GRPO~\cite{shao2024deepseekmath}, as our basic reinforcement learning method. However, we modify the reward function to make the GRPO have a more stable process and one that is suitable for manipulating the browser.
\section{Method}
In this section, we present a detailed description of the proposed method. Firstly, we introduce an automatic process for generating verifiable tasks to manipulate web browsers and other computer applications. Subsequently, we elaborate on the sample purification method leveraging LLM-as-judge; this approach ensures the data exhibits a favorable difficulty distribution, which facilitates effective model learning. 
Following this, we provide a detailed explanation of the proposed reinforcement learning (RL) method. Specifically, we first review the GRPO algorithm, then describe how we adapt it for training our model through modifications to the learning process and reward function. Next, we introduce the Grounding-Based Generalization Enhancement module, which leverages the model’s grounding capability to enhance its planning performance—thereby enabling the model to effectively manipulate target applications.
\subsection{Data Engine for Self-evolution}
One of the key factors driving model evolution is the ability to scale training data, as this directly enhances the model’s capacity for exploration. However, collecting human-like computer operation data—data that mimics real human interaction with computers—is highly complex, leading to insufficient volumes that hinder effective data scaling.
To address this limitation, we propose a novel approach for generating a trajectory dataset of operations on diverse computer software. Specifically, this approach comprises two core components: (1) Generation of closed-loop verifiable tasks; (2) Generation and Evaluation of Trajectory Extraction.

\subsubsection{Generation of closed-loop verifiable tasks}
\begin{figure}
    \centering
    \includegraphics[width=\linewidth]{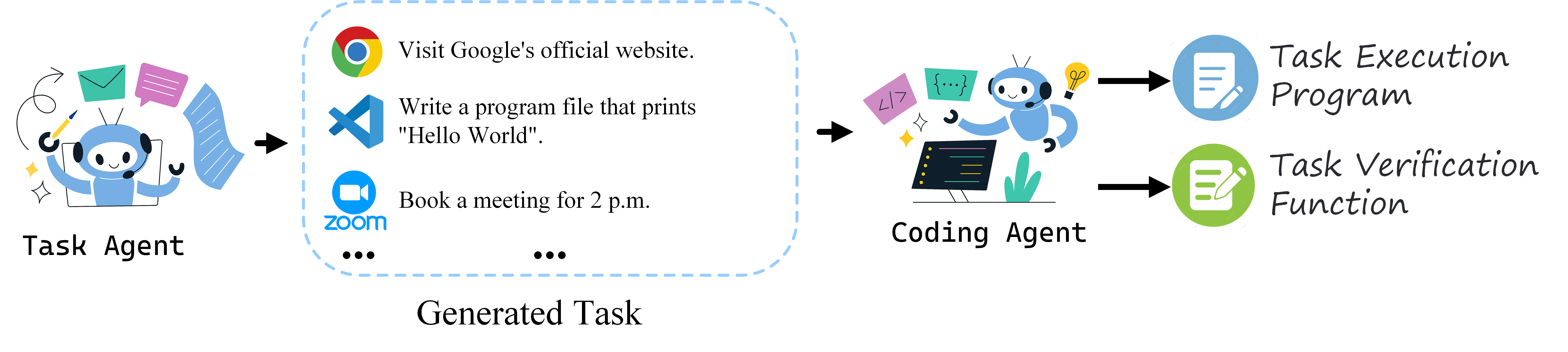}
    \caption{Illustration of the closed-loop task data generation pipeline. The pipeline consists of two core agents: (1) \textit{Task Agent}: Generates task instructions (using few-shot examples of real software tasks) and checks executability/duplication; (2) \textit{Coding Agent}: Takes task instructions and guidelines as input to synthesize Python-based execution programs (for task completion) and verification programs (for validating task success). Only tasks passing automatic execution and verification are retained.}
    \label{fig:data_gen}
    \vspace{-5mm}
\end{figure}

To enable the agent to better handle complex multimodal tasks and dynamic environment interactions, we design a closed-loop task data generation pipeline, as illustrated in Figure~\ref{fig:data_gen}. 

Firstly, the proposed pipeline will use a Task Agent to produce many task instructions. The task agent will use some actual tasks of a specific software as the few-shot examples to evoke the agent's in-context learning of the target sofware. During the generation process, the task agent will periodically check whether the generated tasks can be executed and verified and whether there are some similar tasks have been generated. Meanwhile, the agent will produce some guidelines of how to complete the task. 
Secondly, the code generation agent will take the generated task instruction and the guideline as inputs to generate a batch of execution and verification program. For execution program, it's target is to achieve tasks' goals described in instructions by using the Python. And for the verification program, it judges whether the execution programs' output matches the tasks' requirement. 
In the end, we will automatically run pairs of execution program and verification to select the correct verification program. So in this way, we can get the verifiable tasks.  

\begin{figure*}
    \centering
    \includegraphics[width=\linewidth]{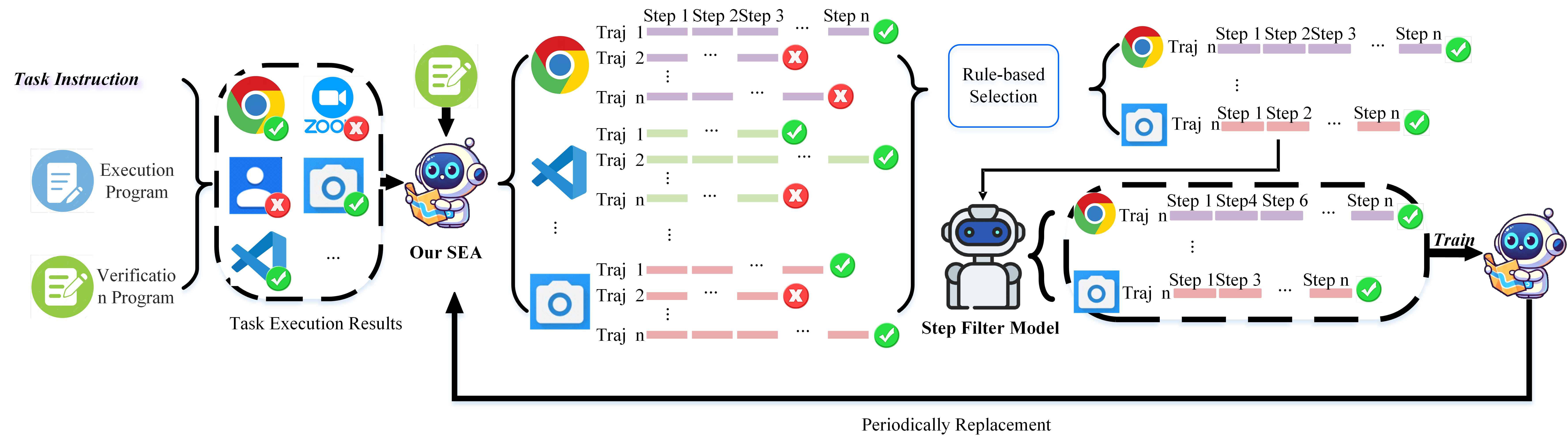}
    \caption{Illustration of the data generation and multi-stage trajectory filtering strategy. This figure illustrates the data generation and multi-stage trajectory filtering strategy. Beginning with task instructions, execution programs, and verification programs, our SEA generates multiple trajectories. These trajectories subsequently go through rule-based selection and filtering by a step filter model. Eventually, the filtered outcomes are utilized to periodically replace and update our SEA for performance refinement.}
    \label{fig:data_filter}
    \vspace{-5mm}
\end{figure*}

\subsubsection{Generation and Assessment for Trajectory Extraction}
To enable the agent to learn the most relevant and effective action sequences, we propose a novel multi-stage trajectory filtering strategy, as illustrated in Figure~\ref{fig:data_filter}. 
After getting the task instruction and verification program, we will firstly use our cold-start model, UI-TARS-1.5-7B~\cite{qin2025ui}, to perform multiple rounds of inference and obtain diverse task trajectories. 
And the verification program is then used to select only those trajectories that lead to successful task completion. 
For each task, we prioritize trajectories with the minimal number of steps based on step-length information. 
Additionally, to further ensure the quality of each step within the trajectory, we train a step filtering model to filter out redundant or erroneous steps, thereby yielding optimal training trajectories. And the step model will be introduced in the next section.
It is notable that during the training process, we will gradually replace the cold-start model with our trained model to generate trajectories for later training stages. In this way, we can make our model evolve through the whole training phase.

\subsubsection{Filtering data with step model}
For the step filtering model, its task is to determine whether the action has been successfully executed based on the two screenshots before and after this step, the thought and action information. 
We manually annotated some ground truth data and continuously adjusted prompts to ensure that the model's performance follows our filtering criteria. Then, we randomly extracted 10,000 step data from the task trajectory and used the Qwen2.5-VL-72B~\cite{bai2025qwen2} model and adjusted the prompt for inference. Each piece of data was subjected to multiple inferences and voting to ensure accuracy as much as possible, and finally used to train the step filtering model. Through this method, we distill the capabilities of the large model into the small model, making the step filtering model lighter while maintaining excellent performance consistent with the large model.
\begin{figure}[h]
    \centering
    \includegraphics[width=\linewidth]{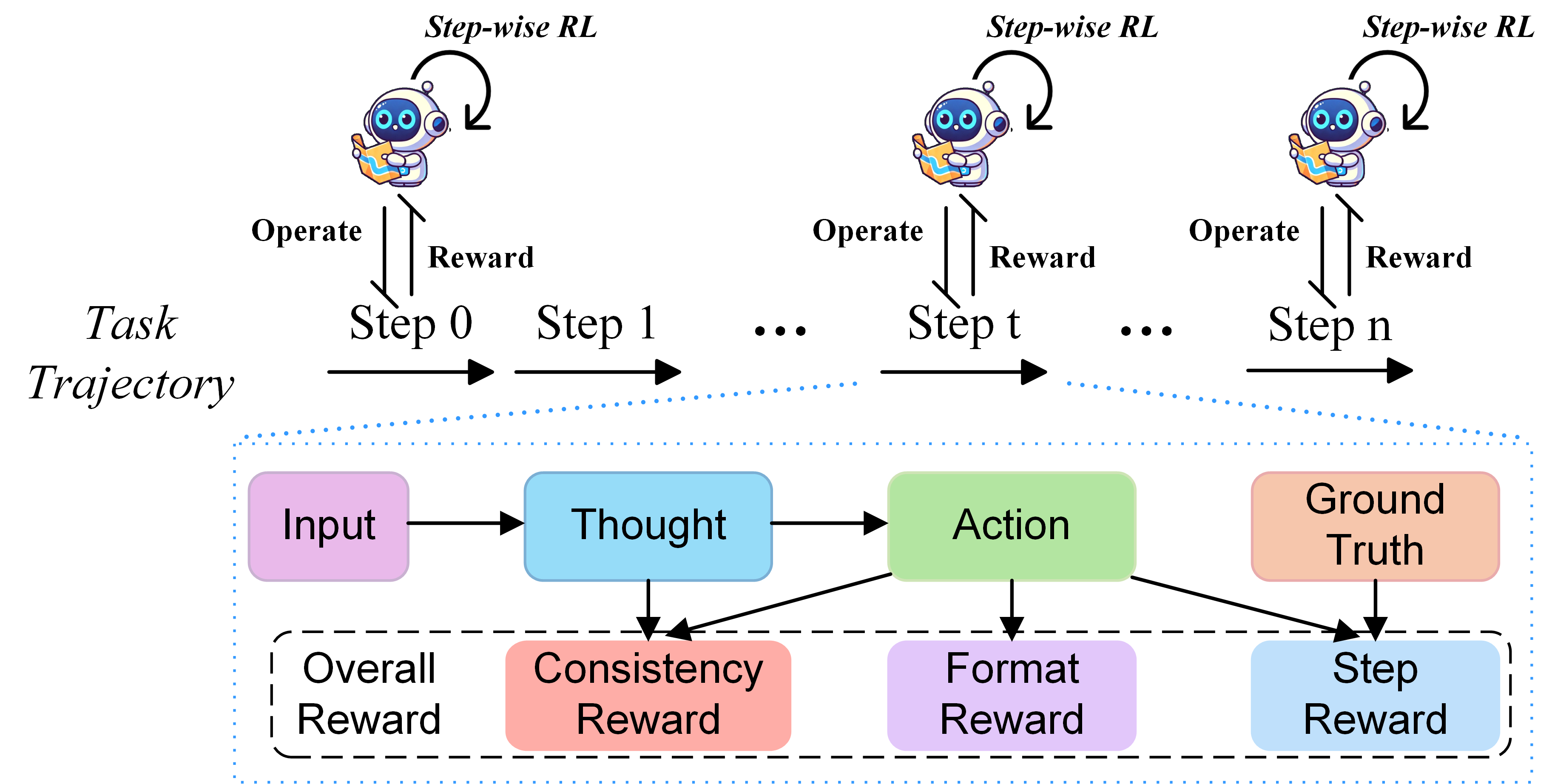}
    \caption{Illustration of Trajectory Reasoning by Step-wise Reinforcement Learning. This figure illustrates the process of trajectory reasoning via step-wise reinforcement learning (RL). Step-wise RL allows an agent to operate and obtain rewards at each step from Step 0 to Step n along a task trajectory. Each step comprises Input, Thought, Action, and Ground Truth components, with rewards calculated based on four facets: Overall Reward, Consistency Reward, Format Reward, and Step Reward.}
    \label{fig:step_level}
    \vspace{-5mm}
\end{figure}
\subsection{Step-wise Reinforcement Learning}

In this part, we will introduce Trajectory Reasoning by Step-wise Reinforcement Learning. 
Specifically, we will use the UI-TARS-7B~\cite{qin2025ui} as our cold-start model, and based on this model, we will use the proposed methods to produce the final agent. 
\subsubsection{Preliminary of GRPO}
Although there are lots of improvements in the GRPO, in this paper, we still choose it, as a result of its balance between performance and tidiness. 

\noindent Group Relative Policy Optimization (GRPO)~\cite{shao2024deepseekmath} is a reinforcement learning method designed for efficient fine-tuning of large language models (LLMs) without relying on an explicit value function or critic. GRPO replaces the trained value function in standard Proximal Policy Optimization (PPO) with the average rewards of multiple outputs for the same query, making it particularly suitable for LLMs.
Given a batch of \( G \) responses \( \{o_i\}_{i=1}^{G} \) from a query \( q \), each consisting of a sequence of tokens \( o_i = (o_i(1), \dots, o_i(T)) \), the GRPO objective is defined as:

\begin{equation}
\begin{aligned}
\mathcal{T}_{\text{GRPO}}(\theta) &= \mathbb{E}\left[q \sim \mathcal{P}(Q),\left\{o_{i}\right\}_{i=1}^{G} \sim \pi_{\theta_{\text{old}}}(O | q)\right] \\
&\quad \frac{1}{G} \sum_{i=1}^{G} \frac{1}{|o_{i}|} \sum_{t=1}^{|o_{i}|} \left\{ \min\left[ \frac{\pi_{\theta}(o_{i,t} | q, o_{i,<t})}{\pi_{\theta_{\text{old}}}(o_{i,t} | q, o_{i,<t})} \hat{A}_{i,t}, \right. \right. \\
&\quad \left. \text{clip}\left( \frac{\pi_{\theta}(o_{i,t} | q, o_{i,<t})}{\pi_{\theta_{\text{old}}}(o_{i,t} | q, o_{i,<t})}, 1-\varepsilon, 1+\varepsilon \right) \hat{A}_{i,t} \right] \\
&\quad \left. -\beta \mathbb{D}_{\text{KL}}\left[ \pi_{\theta} \parallel \pi_{\text{ref}} \right] \right\}
\end{aligned}
\label{equ_GPRO}
\end{equation}

where $\mathbb{D}_{\text{KL}}[\pi_{\theta} \parallel \pi_{\text{ref}}]$ denotes the Kullback-Leibler divergence between the current policy $\pi_{\theta}$ and the reference policy $\pi_{\text{ref}}$, defined in Equation~\ref{equ_GPRO}:
\begin{equation}
\mathbb{D}_{\text{KL}}\left[ \pi_{\theta} \parallel \pi_{\text{ref}} \right] = \frac{\pi_{\text{ref}}(o_{i,t} | q, o_{i,<t})}{\pi_{\theta}(o_{i,t} | q, o_{i,<t})} - \log \frac{\pi_{\text{ref}}(o_{i,t} | q, o_{i,<t})}{\pi_{\theta}(o_{i,t} | q, o_{i,<t})} - 1
\end{equation}

$\hat{A}_{i,t}$ represents the group-normalized advantage for token $t$ in response $o_{i}$, calculated as:
\begin{equation}
\hat{A}_{i,t} = \frac{r_{i} - \mu}{\sigma}
\end{equation}

Where the $r_{i}$ is the total reward of $o_{i}$, while $\mu$ and $\sigma$ are the mean and standard deviation of rewards in the group.

\subsubsection{Step-wise RL training}

Unlike single-step reinforcement learning, computer use agents are required to perform multi-step reasoning and decision-making while interacting with dynamic environments that provide visual feedback. However, training on long-horizon tasks often suffers from the problem of sparse rewards. To address this issue, we propose a step-wise training method, as illustrated in Figure~\ref{fig:step_level}. 
In our training setup, each task trajectory consists of multiple steps, where each step comprises a screen observation $s_i$, a keyboard or mouse action $a_i$, and an associated reward $r_i$, with $i$ denoting any step in the trajectory.

\subsubsection{Rewards}
To predict the next action $a_{i+1}$, the model takes as input the user's request (i.e., task goal), the current screen observation, and the historical sequence of actions and states. As shown in the ~\ref{fig:step_level}, we have used multiple rewards, specifically, we introduce three types of reward mechanisms to optimize the training process:

\noindent\textbf{Step Reward.} For each step in a trajectory, we define a step-level reward based on whether it succeeds. If the agent completes a step and the execution result matches the ground truth, the reward is set to $r_i = 1$; otherwise, $r_i = 0$. This step-wise reward structure is particularly beneficial for long-horizon tasks, as it allows the agent to learn effective action sequences over multiple interactions. By providing clear and immediate feedback at each step, the agent can better identify which actions are effective and which require adjustment.

\noindent\textbf{Reasoning and Action Consistency Reward.} During training, every action produced by the agent is accompanied by a corresponding thought process. If the predicted action is consistent with the reasoning (i.e., aligned in logic and intent), a consistency reward $r_c = 1$ is given; otherwise, $r_c = 0$. This mechanism effectively constrains the model to maintain alignment between its reasoning and the resulting action, leading to more coherent and reliable behavior.

\noindent\textbf{Action Format Reward.} During each rollout, responses from the agent are parsed into discrete actions. If a response fails to conform to the required action schema and cannot be parsed, we assign a penalty of $r_f = 0$, otherwise $r_f = 1$. This encourages the model to generate syntactically valid and executable actions.


\subsubsection{Training Objective} We divide each task trajectory $\tau = (s_0, a_0, \allowbreak \ldots, \allowbreak s_n, a_n)$ into multiple steps. At each step, the agent observes the screen state $s_i$ and generates an action $a_i$ to fulfill the task instruction $x_i \in D$. Our training objective is to maximize the expected reward for step completion, consistency between thoughts and actions, and action format:
\begin{equation}
    \max_{\theta} \mathbb{E}_{x_i \sim D, \tau_i \sim \pi_{\theta}} [r_i(x_i, \tau_i) + r_c(x_i, \tau_i) + r_f(x_i, \tau_i)]
\label{equ_train}
\end{equation}

We optimize this objective using GRPO, which estimates token-level advantages via group-normalized trajectory rewards, allowing efficient and scalable training without a value function.

\begin{table*}[h]
\centering
\caption{Comparison of Various Planners and Grounding Methods on ScreenSpot-Pro. It presents a comparison of various planners and grounding methods in terms of their performance on the ScreenSpot-Pro dataset. It reports results across different task categories (Development, Creative, CAD, Scientific, Office, and OS) for both Text and Icon modalities, along with the average (Avg) scores for each agent model.}
\label{tab:grounding_result}
\resizebox{\textwidth}{!}{%
\begin{tabular}{lccccccccccccccccccccc}
\toprule
\multirow{2}{*}{\textbf{Agent Model}} & \multicolumn{3}{c}{\textbf{Development}} & \multicolumn{3}{c}{\textbf{Creative}} & \multicolumn{3}{c}{\textbf{CAD}} & \multicolumn{3}{c}{\textbf{Scientific}} & \multicolumn{3}{c}{\textbf{Office}} & \multicolumn{3}{c}{\textbf{OS}} & \multicolumn{3}{c}{\textbf{Avg}} \\
\cmidrule(lr){2-4} \cmidrule(lr){5-7} \cmidrule(lr){8-10} \cmidrule(lr){11-13} \cmidrule(lr){14-16} \cmidrule(lr){17-19} \cmidrule(lr){20-22}
 & Text & Icon & Avg & Text & Icon & Avg & Text & Icon & Avg & Text & Icon & Avg & Text & Icon & Avg & Text & Icon & Avg &  Text & Icon & Avg\\
\midrule
QwenVL-7B~\cite{bai2023qwenvlversatilevisionlanguagemodel} & 0.0 & 0.0 & 0.0 & 0.0 & 0.0 & 0.0 & 0.0 & 0.0 & 0.0 & 0.7 & 0.0 & 0.4 & 0.0 & 0.0 & 0.0 & 0.0 & 0.0 & 0.0 & 0.1 & 0.0 & 0.1 \\
GPT-4o~\cite{hurst2024gpt} & 1.3 & 0.0 & 0.7 & 1.0 & 0.0 & 0.6 & 2.0 & 0.0 & 1.5 & 2.1 & 0.0 & 1.2 & 1.1 & 0.0 & 0.9 & 0.0 & 0.0 & 0.0 & 1.3 & 0.0 & 0.8 \\
SeeClick~\cite{cheng2024seeclick} & 0.6 & 0.0 & 0.3 & 1.0 & 0.0 & 0.6 & 2.5 & 0.0 & 1.9 & 3.5 & 0.0 & 2.0 & 1.1 & 0.0 & 0.9 & 2.5 & 0.0 & 1.5 & 1.8 & 0.0 & 1.1 \\
Qwen2-VL-7B~\cite{wang2024qwen2} & 2.6 & 0.0 & 1.3 & 1.5 & 0.0 & 0.9 & 0.5 & 0.0 & 0.4 & 6.3 & 0.0 & 3.5 & 3.4 & 1.9 & 3.0 & 0.9 & 0.0 & 0.5 & 2.5 & 0.2 & 1.6 \\
OS-Atlas-4B~\cite{wu2024atlas} & 7.1 & 0.0 & 3.7 & 3.0 & 1.4 & 2.3 & 2.0 & 0.0 & 1.5 & 9.0 & 5.5 & 7.5 & 5.1 & 3.8 & 4.8 & 5.6 & 0.0 & 3.1 & 5.0 & 1.7 & 3.7 \\
ShowUI-2B~\cite{lin2025showui} & 16.9 & 1.4 & 9.4 & 9.1 & 0.0 & 5.3 & 2.5 & 0.0 & 1.9 & 13.2 & 7.3 & 10.6 & 15.3 & 7.5 & 13.5 & 10.3 & 2.2 & 6.6 & 10.8 & 2.6 & 7.7 \\
CogAgent-18B~\cite{hong2024cogagent} & 14.9 & 0.7 & 8.0 & 9.6 & 0.0 & 5.6 & 7.1 & 3.1 & 6.1 & 22.2 & 1.8 & 13.4 & 13.0 & 0.0 & 10.0 & 5.6 & 0.0 & 3.1 & 12.0 & 0.8 & 7.7 \\
Aria-UI~\cite{yang2024aria} & 16.2 & 0.0 & 8.4 & 23.7 & 2.1 & 14.7 & 7.6 & 1.6 & 6.1 & 27.1 & 6.4 & 18.1 & 20.3 & 1.9 & 16.1 & 4.7 & 0.0 & 2.6 & 17.1 & 2.0 & 11.3 \\
UGround-7B~\cite{gou2024navigating} & 26.6 & 2.1 & 14.7 & 27.3 & 2.8 & 17.0 & 14.2 & 1.6 & 11.1 & 31.9 & 2.7 & 19.3 & 31.6 & 11.3 & 27.0 & 17.8 & 0.0 & 9.7 & 25.0 & 2.8 & 16.5 \\
Claude Computer Use~\cite{hu2024dawn} & 22.0 & 3.9 & 12.6 & 25.9 & 3.4 & 16.8 & 14.5 & 3.7 & 11.9 & 33.9 & 15.8 & 25.8 & 30.1 & 16.3 & 26.9 & 11.0 & 4.5 & 8.1 & 23.4 & 7.1 & 17.1 \\
OS-Atlas-7B~\cite{wu2024atlas} & 33.1 & 1.4 & 17.7 & 28.8 & 2.8 & 17.9 & 12.2 & 4.7 & 10.3 & 37.5 & 7.3 & 24.4 & 33.9 & 5.7 & 27.4 & 27.1 & 4.5 & 16.8 & 28.1 & 4.0 & 18.9 \\
UGround-V1-7B~\cite{gou2024navigating} & --  & -- & 35.5 & -- & -- & 27.8 & -- & -- & 13.5 & -- & -- & 38.8 & -- & -- & 48.8 & -- & -- & 26.1 & -- & -- & 31.1 \\
UI-TARS-2B~\cite{qin2025ui} & 47.4 & 4.1 & 26.4 & 42.9 & 6.3 & 27.6 & 17.8 & 4.7 & 14.6 & 56.9 & 17.3 & 39.8 & 50.3 & 17.0 & 42.6 & 21.5 & 5.6 & 14.3 & 39.6 & 8.4 & 27.7 \\
UI-TARS-7B~\cite{qin2025ui} & 58.4 & 12.4 & 36.1 & 50.0 & 9.1 & 32.8 & 20.8 & 9.4 & 18.0 & 63.9 & \textbf{31.8} & 50.0 & 63.3 & 20.8 & 53.5 & 30.8 & 16.9 & 24.5 & 47.8 & 16.2 & 35.7 \\
UI-TARS-72B~\cite{qin2025ui} & 63.0 & 17.3 & 40.8 & 57.1 & 15.4 & 39.6 & 18.8 & 12.5 & 17.2 & 64.6 & 20.9 & 45.7 & 63.3 & 26.4 & 54.8 & 42.1 & 15.7 & 30.1 & 50.9 & 17.5 & 38.1 \\
\textbf{Ours(Grounding Model)} & \textbf{76.0} & \textbf{21.4} & \textbf{49.5} & \textbf{61.6} & \textbf{19.6} & \textbf{44.0}  & \textbf{45.2} & \textbf{18.8} & \textbf{38.7} & \textbf{80.6} & \textbf{31.8} & \textbf{59.4} & \textbf{84.2} & \textbf{54.7} & \textbf{77.4} & \textbf{57.9} & \textbf{33.7} & \textbf{46.9} & \textbf{67.0} & \textbf{27.3} & \textbf{51.9} \\
\bottomrule
\end{tabular}
}
\end{table*}

\begin{table}[]
\caption{Grounding results on the ScreenSpot-V2. It presents the grounding results of various models on the ScreenSpot-V2 dataset, including their performance in Text and Icon modalities across three platforms (Desktop, Mobile, Web) as well as the average (AVG) scores.}
\small
\label{tab:ssp2_results}
\resizebox{0.5\textwidth}{!}{%
\begin{tabular}{llllllll}
\hline
\multicolumn{1}{c}{\multirow{2}{*}{MODEL}} & \multicolumn{2}{c}{Desktop} & \multicolumn{2}{c}{Mobile} & \multicolumn{2}{c}{Web} & \multirow{2}{*}{AVG} \\ \cline{2-7}
\multicolumn{1}{c}{}   & Text & Icon & Text  & Icon & Text & Icon &      \\ \hline
UI-Venus-72B~\cite{gu2025ui}           & 95.9 & 90.0 & 99.7  & 93.8 & 96.2 & 92.6 & 95.3 \\
Holo1.5-72B~\cite{hai2025holo15modelfamily}            & 96.1 & 86.9 & 100.0 & 92.7 & 95.9 & 90.7 & 94.4 \\
UI-Venus-7B~\cite{gu2025ui}            & 96.9 & 90.7 & 99.0  & 90.0 & 96.2 & 88.7 & 94.1 \\
Holo1.5-7B~\cite{hai2025holo15modelfamily}             & 96.7 & 88.9 & 99.2  & 91.1 & 95.4 & 86.1 & 93.3 \\
GUI-ARP-7B~\cite{ye2025gui}             & 97.2 & 85.7 & 96.5  & 89.0 & 94.3 & 84.8 & 91.8 \\
Holo1.5-3B~\cite{hai2025holo15modelfamily}             & 95.0 & 89.7 & 99.2  & 88.0 & 91.8 & 84.8 & 91.7 \\
Qwen2.5-VL-7B-Instruct~\cite{bai2025qwen2} & 87.6 & 65.7 & 99.0  & 84.4 & 90.2 & 79.8 & 86.5 \\
AGUVIS-7B~\cite{xu2024aguvis}              & 93.3 & 77.9 & 95.5  & 81.5 & 91.0 & 77.8 & 87.3 \\
OSAtlas-7B~\cite{wu2024atlas}             & 90.7 & 63.6 & 93.8  & 73.9 & 89.7 & 77.3 & 83.3 \\
ZonUI-3B~\cite{hsieh2025zonui}               & 92.3 & 74.3 & 98.6  & 82.9 & 88.0 & 74.4 & 86.6 \\
UGround-7B~\cite{gou2024uground}           & 85.1 & 61.4 & 84.5  & 61.6 & 84.6 & 71.9 & 76.3 \\
ShowUI-2B~\cite{lin2025showui}            & 78.9 & 59.3 & 92.1  & 75.4 & 84.2 & 61.1 & 77.3 \\
OSAtlas-4B~\cite{wu2024atlas}             & 64.4 & 46.4 & 82.8  & 64.0 & 78.6 & 60.1 & 68.5 \\
CogAgent-18B~\cite{hong2024cogagent}         & 75.8 & 20.7 & 69.3  & 27.0 & 74.4 & 31.5 & 52.8 \\
MiniCPM-V-7B~\cite{yao2024minicpm}         & 69.1 & 18.6 & 59.7  & 18.0 & 67.9 & 26.6 & 45.9 \\
SeeClick-7B~\cite{cheng2024seeclick}          & 69.6 & 30.7 & 77.9  & 48.3 & 57.3 & 22.2 & 53.9 \\
Qwen2.5-VL-3B-Instruct~\cite{bai2025qwen2} & 54.1 & 30.0 & 62.1  & 46.4 & 31.2 & 48.3 & 46.9 \\
GPT-4o~\cite{hurst2024gpt}                 & 24.2 & 19.3 & 26.6  & 24.2 & 12.8 & 11.8 & 20.1 \\ \hline
\textbf{Ours }                  & 98.3 & 84.8 & 94.3  & 77.1 & 91.0 & 84.7 & 89.6 \\ \hline
\end{tabular}
}
\end{table}

\begin{table*}[htbp]
\centering
\caption{Results with max steps of 50 on the online OS-World Benchmark. We have submitted our final model to the OSWorld website and compared with other SOTA methods, and our method achieves the best performance with a small number of parameters.}
\label{tab:osw_online_results}
\small 
\tabcolsep=3pt 
\renewcommand{\arraystretch}{1.1} 
\newcolumntype{d}[1]{D{.}{.}{#1}} 

\resizebox{\linewidth}{!}{
\begin{tabularx}{\linewidth}{
  l 
  l 
  d{3.1} 
  d{5.2} 
  *{10}{d{4.2}} 
}
\toprule
\textbf{Category} 
& \textbf{Model Name} 
& \multicolumn{1}{c}{\textbf{Succ Rate}} 
& \multicolumn{1}{c}{\textbf{Chrome}} 
& \multicolumn{1}{c}{\textbf{Gimp}} 
& \multicolumn{1}{c}{\makecell{Libre\\Calc}} 
& \multicolumn{1}{c}{\makecell{Libre\\Impress}} 
& \multicolumn{1}{c}{\makecell{Libre\\Writer}} 
& \multicolumn{1}{c}{\makecell{Multi\\Apps}} 
& \multicolumn{1}{c}{OS} 
& \multicolumn{1}{c}{\makecell{Thunder\\bird}} 
& \multicolumn{1}{c}{VLC} 
& \multicolumn{1}{c}{\makecell{VS\\Code}} 
\\
\midrule
\multirow{4}{*}{\makecell{Closed Source\\Model}} 
& claude-4-sonnet-20250514   & 43.9 & 24.96 & 13.00 & 15.00 & 21.96 & 14.00 & 26.50 & 11.00 & 11.00 & 7.00 & 14.00 \\
& claude-3-7-sonnet-20250219 & 35.8 & 23.96 & 10.00 & 15.00 & 16.96 & 10.00 & 16.42 & 12.00 & 8.00  & 4.00 & 13.00 \\
& o3                         & 17.2 & 10.00 & 10.00 & 4.00  & 2.00  & 5.00  & 11.00 & 9.00  & 3.00  & 2.99 & 5.00  \\
& computer-use-preview       & 31.3 & 16.96 & 9.00  & 7.00  & 13.96 & 6.00  & 14.70 & 17.00 & 10.00 & 2.00 & 16.00 \\
\midrule
\multirow{2}{*}{\makecell{Large\\Params}} 
& opencua-32b~\cite{wang2025opencuaopenfoundationscomputeruse}                & 34.1 & 18.96 & 18.00 & 7.00  & 15.99 & 10.00 & 12.35 & 12.00 & 8.00  & 5.00 & 13.00 \\
& uitars-72b-dpo~\cite{qin2025ui}             & 25.8 & 14.96 & 16.00 & 6.00  & 11.96 & 10.00 & 6.24  & 8.00  & 5.00  & 4.00 & 11.00 \\
\midrule
\multirow{5}{*}{\makecell{Small\\Params}} 
& opencua-7b~\cite{wang2025opencuaopenfoundationscomputeruse}                 & 28.2 & 17.96 & 10.00 & 5.00  & 16.96 & 8.00  & 8.69  & 10.00 & 6.00  & 5.44 & 11.00 \\
& uitars-1.5-7b~\cite{ui-tars-15-seed}              & 27.3 & 12.96 & 13.00 & 2.00  & 16.99 & 9.00  & 9.09  & 6.00  & 7.00  & 3.00 & 11.00 \\
& opencua-qwen2-7b~\cite{wang2025opencuaopenfoundationscomputeruse}           & 20.6 & 14.96 & 6.00  & 3.00  & 12.00 & 10.00 & 4.00  & 2.00  & 7.00  & 5.00 & 10.00 \\
& opencua-a3b~\cite{wang2025opencuaopenfoundationscomputeruse}                & 19.9 & 9.92  & 14.00 & 1.00  & 10.96 & 8.00  & 6.07  & 9.00  & 1.00  & 2.00 & 10.00 \\
& Ours                       & 29.8 & 18.96 & 15.00 & 3.00  & 17.99 & 13.00 & 5.58  & 8.00  & 6.00  & 5.83 & 16.00 \\
\bottomrule
\end{tabularx}%
}
\end{table*}

\subsection{Grounding-Based Generalization Enhancement.}  
After using the step-wise RL method to train the cold-start model using the verifiable tasks collected through the proposed data engine, we can get a model with strong planning ability. However, the computer use agent not only needs the planning also needs the grounding to locate target elements in the screen.  In order to enhance the model's perception ability, we firstly train a grounding model and then merge the planning model with the grounding model. In the end, we use the Temporal Compressed Sensing Mechanism to make agent pay more attention of images. 
\subsubsection{Train Grounding Model}
We will use the supervised fine-tuning to get a grounding model, and the training data is one of the key points influencing its performance. At present, there are numerous open-source datasets in the field of grounding. 

Aguvis~\cite{xu2024aguvis} and OS-Atlas\cite{wu2024atlas} have organized and filtered many previous open-source datasets with high quality, so we have conducted our data filtering based on them. We select 30000 pieces of data from each subset of these two datasets (if fewer than 30000 pieces, we take all), and these 30000 pieces of data should come from different screenshots, rather than other elements of the same screenshot. 
We used these different datasets and the same configuration for SFT training, and evaluated them using screenspot-v2. Finally, we selected five higher-quality subsets from two large datasets as the training set for the grounding model. Through these experiments, we also found that the quality of the dataset is far more critical than the quantity. 

In addition, detailed instructions are helpful for the grounding task. To reconstruct the data samples, we retain the first two steps from \cite{wu2025smoothing}
For sample construction, we retain the first two steps from \cite{wu2025smoothing} while optimizing the third step. The original method discards samples with imprecise predicted coordinates during sample construction. However, we observe that these discarded samples often correspond to more complex tasks, involving richer operational logic and reasoning paths. Therefore, we preserve the original task instructions (the original commands) of these samples and reincorporate them into the training set. To further enhance data diversity and model robustness, we apply data augmentation techniques such as image scaling and color inversion to the filtered dataset.

Finally, we train our grounding model using SFT and GRPO reinforcement learning, employing the following IoU-based loss or reward function:
\begin{equation}
\begin{aligned}
\text{Reward} = \min\Bigg( 1, \ &\text{IoU}(B_p, B_{gt}) \cdot\\ \
&  \frac{1}{\mathbb{I}\left[\text{IoU}(B_p, B_{gt}) \leq 0.7\right] + \varepsilon} \Bigg)
\end{aligned}
\end{equation}


Here, $\mathbb{I}$ denotes the indicator function, which equals 1 if the condition inside the parentheses is satisfied, and 0 otherwise.
$\text{IoU}(B_p, B_{gt}) = \frac{|B_p \cap B_{gt}|}{|B_p \cup B_{gt}|}$ represents the Intersection over Union (IoU) score between the predicted bounding box $B_p$ and the ground truth box $B_{gt}$.
$\varepsilon$ is a small constant (e.g., $10^{-6}$) used to prevent division by zero.
\subsubsection{Model Merge}After getting the grounding model, we need to find a simple and effective way to combine the planning and grounding abilities into one model. Meanwhile, we do not want to use extra computation and data to distill these two models into one. As a result of that, we leverage the DARE~\cite{yu2024language} to merge these two models. Based on the idea of task vectors, this method sets most incremental parameters (i.e., the differences between parameters of the grounding model and planning model) to zero without affecting the performance of original models. DARE randomly discards incremental parameters at a rate of $p$ and rescales the remaining parameters by $1/(1 - p)$ to approximate the original embedding. 
\subsubsection{Temporal Compressed Sensing Mechanism}
To improve training efficiency and reduce computational cost, we propose the TCSM (Temporal Compressed Sensing Mechanism). Specifically, TCSM first differentiates tokens based on their recency. Newer tokens typically carry more information about the current task state and objectives, and thus are assigned higher priority during training. Tokens with relatively lower priority undergo structured compression to reduce computational overhead. 
\begin{equation}
    \mathcal{I}_{\text{TCSM}} = \left\{ \operatorname{Pad}\left( \operatorname{Resize}(I_i) \right) \ \middle|\  i \in \mathcal{H},\ i \geq n - k + 1 \right\}
\end{equation}
Here, $\mathcal{H}$ denotes the sequence of historical images, containing $n$ frames in total. $I_i$ represents the $i$-th frame in the history. To focus on the key temporal context, we retain only the most recent $k$ frames, while the remaining earlier frames are compressed for efficient representation.
\section{Experiments}
In this section, extensive experiments are conducted to compare the proposed agent with other state-of-the-art methods, followed by some relevant analysis of our method.

\subsection{Experimental Setup}
\noindent\textbf{Implementation Details.} For the training part, we use the LLaMa-Factory to train our grounding model, which is used in the model integration. And for the whole model, we use the Easy-R1 to train the model by using the GRPO. During the entire training phase, we use one server with 8 A100 (80GB) GPUs. During the inference phase, we reference the GTA-1~\cite{yang2025gta1} to have better results. Firstly, we used our agent to get $N$ possible results; in our experiment, $N$ equals 8. And then we will use another model to select the best one from them, which differs from GTA-1~\cite{yang2025gta1} in that it utilizes GPT-o3.

\noindent \textbf{Datasets and Metrics.} As we have mentioned before, we will use the generated verifiable tasks to produce the trajectory training data. In our experiments, we use the 400 tasks. And for each inference in the trajectory, we produce eight results. For the evaluation, we evaluate our improved grounding model on the ScreenSpot-V2~\cite{cheng2024seeclick, wu2024atlas} and ScreenSpotPro~\cite{li2025screenspot} datasets using accuracy as the evaluation metric. Subsequently, we evaluate our proposed model on the OSWorld benchmark~\cite{xie2024osworld}, which is designed to evaluate the computer use agent's performance on open-ended GUI tasks. This benchmark comprises 369 tasks spanning real-world web and desktop applications, providing a diverse and challenging platform to evaluate the agent’s ability to complete user tasks within a Linux environment. Performance is measured primarily by task success rate.

\subsection{Inference Stage}
During the inference stage, we build upon the GTA1~\cite{yang2025gta1} and replace the original inference model with our proposed model SEA. As a result, both the sampling and selection of candidate trajectories during inference are handled entirely by SEA, without relying on any external scoring models. This design significantly reduces computational overhead and effectively improves decision-making efficiency.

\subsection{Experimental Results}
\subsubsection{Grounding Performance}
We first evaluate the performance of our improved and trained grounding model. Among models of comparable scale, our grounding model achieves SOTA performance, as shown in ~\cref{tab:grounding_result}. As shown in ~\cref{tab:grounding_result}, our grounding model has each domain in the ScreenSpotPro~\cite{li2025screenspot} and outperforms the larger models such as CogAgent-18B~\cite{hong2024cogagent}, Claude Computer Use~\cite{hu2024dawn} and UI-TARS-72B~\cite{qin2025ui}. As ScreenSpotPro~\cite{li2025screenspot} does not explicitly reflect our model's performance in the web domain, we also evaluate our model in the ScreenSpot-V2~\cite{cheng2024seeclick, wu2024atlas}, shown in the ~\cref{tab:ssp2_results}.

\subsubsection{OS-World Performance}
We compare our method with state-of-the-art approaches on the OSWorld benchmark, with results summarized in~\cref{tab:osw_online_results}. As shown, despite utilizing significantly fewer parameters and incurring lower training costs, the proposed method still surpasses the larger model UI-TARS-72B-DPO~\cite{qin2025ui} by 3.0\% in task success rate. Furthermore, we reproduced the baseline model UI-TARS-1.5-7B~\cite{ui-tars-15-seed} under the same experimental conditions. Compared to this baseline, SEA achieved a performance improvement of 5.2\% and attained the highest task success rate among models of the same scale on the OSWorld benchmark, outperforming all existing state-of-the-art approaches. These results demonstrate the effectiveness of the proposed method in handling complex and realistic user tasks across diverse scenarios, highlighting the robustness and generalization capability of our approach. Meanwhile, as shown in this table, the model without the enhancement also achieves the SOTA performance, which proves the effectiveness of the data generation of verifiable tasks and the step-wise reinforcement learning.
\begin{table}[]
\caption{The offline results for the ablation study. In order to conveniently compare, we have evaluated some SOTA methods on our deployed OSWorld environment. We choose some representative methods, and the UI-TARS-1.5-7B is our foundation model.}
\label{tab:offline_results}
\begin{tabular}{lc}
\hline
Model                                                       & OSWorld (\%) \\ \hline
OpenCUA-3B~\cite{wang2025opencuaopenfoundationscomputeruse}         & 18.8         \\
OpenCUA-Qwen2-7B~\cite{wang2025opencuaopenfoundationscomputeruse}   & 23.1         \\
OpenCUA-7B~\cite{wang2025opencuaopenfoundationscomputeruse}         & 26.6         \\
UI-TARS-72B-DPO~\cite{qin2025ui}      & 27.1         \\
UI-TARS-1.5-7B~\cite{ui-tars-15-seed} & 24.9         \\ \hline
Ours(w/o merging + w/o enhancement)                         & 27.6         \\
Ours (w/o enhancement)                                      & 28.1         \\
Ours                                                        & 30.1         \\ \hline
\end{tabular}
\end{table}

\section{Ablation Study}
In this section, we will figure out the function of modules in the proposed method through experiments. Firstly, we will demonstrate the effect of merging the grounding and planning models, and secondly, we will show whether the Temporal Compressed Sensing Mechanism enhances the model's performance.
\subsection{Merging models}
As we have mentioned before, we have trained another grounding model to enhance the model's ability to recognize screenshot information. Different from the previous methods using the two-stage training or distill models, we use the model merging method to make the final model have both the ability to extract information from the desktop and plan. As shown in ~\cref{tab:offline_results}, before merging, our model achieves 27.6 in our offline evaluation. The reason why we use the offline instead of submitting the ablation experiment to the official online test is that we need to do many evaluations during the whole process. Meanwhile, we have submitted the final model to the official website and the result is shown in ~\cref{tab:osw_online_results}. In order to have a comparable baseline, we have test the other SOTA methods in our offline evaluation. Comparing the baseline, the merging can improve our model's accuracy from $27.6~\%$ to $28.1~\%$ for operating computers. We believe that the merging of grounding can enhance the planning model's ability to understand screen images. During the process of using the proposed reinforcement learning to train models, the distribution of weights shifts to improve their ability to reason. So, when we merge the grounding model's weights from layer to layer, it can strike a balance between reasoning and understanding the desktop images, and in other words, achieve better performance in the overall metric. This indicates that by integrating weights from different task modules and adopting a self-evolution mechanism, the model can achieve better generalization to novel task scenarios. We believe that this type of enhancement enables the original model to gain a deeper understanding of the PC's screenshots, which will help the model produce more precise plans for various software, including new applications.

\subsection{Generalization Enhancement on planning.} 
Besides the merging process, we also proposed using the Temporal Compressed Sensing Mechanism(TCSM) to enhance the final model. 
During our experiments, we found that TCSM is a simple but effective method to improve the model's performance. As shown in ~\cref{tab:offline_results}, the TCSM can improve the model from 28.1 to 30.1 without extra training and inference cost. We think the reason why it can improve the performance is that the temporal compression of the image is a kind of compression of the historical observation space, which can make the model focus on the current information and invoke the model's reasoning more. Although the history, including the previous actions and the corresponding results, can make the model have the knowledge of what has happened, the superfluous information will hinder the model from making a correct judgment. Nevertheless, the TCSM can strike a balance between using historical knowledge and reasoning.

\section{Conclusion} 

In this paper, we propose a suite of innovations to develop the \textit{Self-Evolution Agent (SEA)} for autonomous computer operation, addressing key challenges in data scalability, long-horizon training, and multi-capability integration. 

First, we design a closed-loop pipeline to generate verifiable task trajectories: a Task Agent generates executable instructions, while a Coding Agent synthesizes execution/verification programs. The GATE method further refines trajectories by prioritizing short, successful paths and filtering redundant steps. Second, we introduce \textit{Trajectory Reasoning by Step-wise Reinforcement Learning (TR-SRL)}, which replaces monolithic long-horizon training with step-wise rewards (step success, thought-action consistency, action format), enabling efficient learning. Third, we propose a \textit{Grounding-Based Generalization Enhancement} method: merging grounding and planning models via DARE and optimizing perception with the Temporal Compressed Sensing Mechanism (TCSM), achieving dual capabilities without extra training.

Experimental results on the OSWorld benchmark show that SEA (7B parameters) outperforms all same-scale models (e.g., 5.2\% improvement over UI-TARS-1.5-7B) and matches the performance of larger models (e.g., UI-TARS-72B-DPO). Future work will focus on extending SEA to complex multi-application scenarios and further reducing training/inference latency for edge-device deployment.
\clearpage
\bibliographystyle{ACM-Reference-Format}
\bibliography{sample-base}










\end{document}